\documentclass[twoside,11pt]{article}

\usepackage[abbrvbib, preprint]{jmlr2e}

\usepackage{booktabs}
\usepackage{amsmath}

\jmlrheading{1}{2021}{1-48}{11/21}{__/__}{____}{Sia Gholami and Mehdi Noori}

\ShortHeadings{Zero-Shot Open-Book QA}{Gholami and Noori}
\firstpageno{1}

\begin{document}

\title{Zero-Shot Open-Book Question Answering}

\author{\name Sia Gholami \email gholami@amazon.com \\
       \name Mehdi Noori \email nmehdi@amazon.com \\
       \addr Amazon Web Services, CA USA}

\editor{__}

\maketitle

\begin{abstract}%

  Open book question answering is a subset of question answering tasks where the system aims to
  find answers in a given set of documents (open-book) and common knowledge about a topic.
  This article proposes a solution for answering natural language questions from a corpus of Amazon Web Services (AWS)
  technical documents with no domain-specific labeled data (zero-shot).
  These questions can have yes-no-none answers, short answers, long answers, or any combination of the above.
  This solution comprises a two-step architecture in which a retriever finds the right document and an extractor
  finds the answers in the retrieved document.
  We are introducing a new test dataset for open-book QA based on real customer questions
  on AWS technical documentation.
  After experimenting with several information retrieval systems and extractor models based on extractive language
  models, the solution attempts to find the yes-no-none answers and text
  answers in the same pass.
  The model is trained on the The Stanford Question Answering Dataset - SQuAD~\citep{rajpurkar2016squad} and
  Natural Questions~\citep{kwiatkowski2019natural} datasets.
  We were able to achieve 49\% F1 and 39\% exact match score (EM) end-to-end with no domain-specific training.

\end{abstract}


\section{Introduction}\label{sec:introduction}

Question answering (QA) has been a major area of research in Artificial Intelligence and Machine Learning
since the early days of computer science~\citep{voorhees1999trec, moldovan2000structure, brill2002analysis, ferrucci2010building}.
The need for a performant open-book QA solution was exacerbated by rapid growth in available information in niche
domains, the growing number of users accessing this information, and the expanding need for more efficient operations.
QA systems are especially useful when a user searches for specific information and does not have the time - or simply
does not want - to peruse all available documentation related to their search to solve the problem at hand.\par

In this article, open-book QA is defined as the task whereby a system (such as a computer software) answers natural
language questions from a set of available documents (open-book).
These questions can have yes-no-none answers, short answers, long answers, or any combination of the above.
In this work, we did not train the system on our domain-specific documents or questions and answers, a technique
called zero-shot learning~\citep{brown2020language}.
The system should be able to perform with a variety of document types and question and answers without training.
We defined this approach as “zero-shot open-book QA”.
The proposed solution is tested on AWS Documentation dataset.
However, as the models within this solution are not trained on the dataset, the solution can be used in other similar
domains such as finance and law. \par

Software technical documentation is a critical piece of the software development life cycle process.
Finding the correct answers for one’s questions can be a tedious and time-consuming process.
Currently, software developers, technical writers, and marketers are required to spend substantial
time writing documents such as technology briefs, web content, white papers, blogs, and reference guides.
Meanwhile, software developers and solution architects have to spend time searching for specific information
they need.
Our approach to QA aims to help them find it faster.\par

Our work’s key contributions are:
\begin{enumerate}
  \item introduce a new dataset in open-book QA,
  \item propose a two-module architecture to find answers without context,
  \item experiment on ready-to-use information retrieval systems,
  \item infer text and yes-no-none answers in a single forward pass once we find the right document.
\end{enumerate}
\par
The rest of the paper is structured as follows: First, related previous work is summarized.
Then the dataset is described.
Next, details on implementing the zero-shot open-book QA pipeline are provided.
In addition, the experiments are explained, and finally the results along with limitations and next steps are presented.

\section{Related work}\label{sec:related-work}

There are a number of datasets in the literature for natural language QA
~\citep{rajpurkar2016squad, joshi2017triviaqa, khashabi2018looking, richardson2013mctest, lai2017race,
  reddy2019coqa, choi2018quac, tafjord2019quarel,mitra2019declarative}, as well several solutions to tackle
these challenges~\citep{seo2016bidirectional, vaswani2017attention, devlin2018bert, he2011summarization, kumar2016ask,
xiong2016dynamic, raffel2019exploring}.
The natural language QA solutions take a question along with a block of text as context and attempts to find the
correct answer to the original question within the context.
Open-book QA solutions take a question along with a set of documents that may contain the answer to the question,
then the solution attempts to find the answer to the original question within the available set of documents.
Open-book QA solutions have been explored by several research teams including Banerjee et al.~\citep{banerjee2019careful},
which performs QA using fine-tuned extractive language models, and the work of Yasunaga et al.~\citep{yasunaga2021qa}, 
which performs QA using GNNs.
\par
In this paper, we propose an approach that differs from the previous body of work as we do not receive the context but
assume that the answer lies in a set of readily available documents (open-book);
In addition, we are not allowed to train our models on the given questions or set of documents (zero-shot).
Our proposed solution attempts to answers questions from a set of documents with no prior training or
fine-tuning (zero-shot open-book question answering).

\section{Data}\label{sec:data}
Real world open-book QA use cases require significant amounts of time, human effort, and cost to access or
generate domain-specific labeled data. 
For our solution, we intentionally did not use any domain-specific labeled data and ran experiments on popular 
QA datasets and pre-trained models. 
We used feedback from customers to generate a set of 100 questions as the test dataset and used QA datasets, explained in
section~\ref{subsec:squad} and~\ref{subsec:natural-questions}, for training.

\subsection{AWS Documentation Dataset}
\label{subsec:aws-documentation-dataset}
Herein, we present the AWS documentation corpus \footnote{https://github.com/siagholami/aws-documentation} ,
an open-book QA dataset, which contains 25,175 documents along with 100 matched questions and answers.
These questions are based on real customer questions on AWS services.
There are two types of answers: text and yes-no-none answers.
Text answers range from a few words to a full paragraph sourced from a continuous series of words in a document
or from different locations within the same document.
Yes-no-none(YNN) answers can be yes, no, or none for cases where the returned result is empty and does not lead to a
binary answer (i.e., yes or no).
All questions in the dataset have a valid answer within the accompanying documents.
Table~\ref{tab:examples} shows a few examples from the dataset.

\begin{table}[h]
  \caption{\footnotesize{Three sample questions from the test dataset }}
  \label{tab:examples}
  \centering
  \begin{tabular}{p{0.4\linewidth} p{0.45\linewidth} p{0.1\linewidth}}
    \toprule
    Question & Text Answer & YNN Answer \\
    \midrule
    What is the maximum number of rows in a dataset in Amazon Forecast? & 1 billion  & None    \\\\
    Can I stop a DB instance that has a read replica? & You can't stop a DB instance that has a read replica. & No     \\\\
    Is AWS IoT Greengrass HIPAA compliant? & Third-party auditors assess the security and compliance of AWS IoT Greengrass as part of multiple AWS compliance programs. These include SOC PCI FedRAMP HIPAA and others. & Yes    \\
    \bottomrule
  \end{tabular}
\end{table}

\subsection{SQuAD Datasets}
\label{subsec:squad}

The Stanford Question Answering Dataset (SQuAD)\footnote{https://rajpurkar.github.io/SQuAD-explorer/}
is a reading comprehension dataset~\citep{rajpurkar2016squad}, including questions created by crowdworkers on Wikipedia articles.
The answers to these questions is a segment of text from reading passages, or the question might be unanswerable.
SQuAD1.1 comprises 100,000 question-answer pairs on more than 500 articles.
SQuAD2.0 adds 50,000 unanswerable questions written adversarially by crowdworkers to look similar to answerable ones.

\subsection{Natural Questions Dataset}
\label{subsec:natural-questions}

The Natural Questions (NQ) dataset \footnote{https://ai.google.com/research/NaturalQuestions}  includes 400,000 
questions and answers created on Wikipedia articles~\citep{kwiatkowski2019natural}.
Questions consist of real queries which answers can be long (a few sentences), short (a few words) if present on
the page, or null if no long or short answer is present.

\section{Approach}\label{sec:approach}
\begin{figure}[h]
  \label{fig:architecture}
  \centering
  \includegraphics[width=\linewidth]{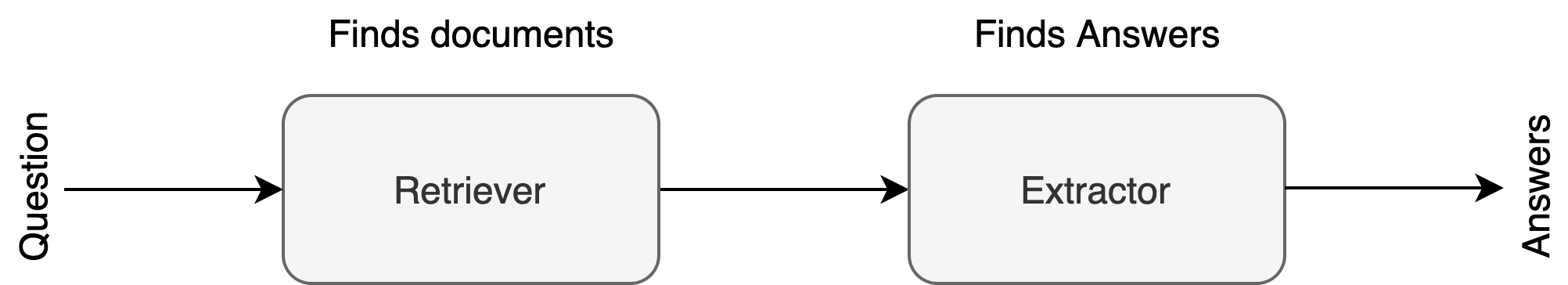}
  \caption{\footnotesize{High level workflow of the solution in one pass}}
\end{figure}

Our approach consists of two high-level modules: retriever and extractor.
Given a question, the retriever tries to find a set of documents that contain the answer;
Then, from these documents, the extractor tries to find the answer.
Figure~\ref{fig:architecture} illustrates a high level workflow of the solution,
and Table~\ref{tab:solution-example} shows an example of the question,retrieved documents, and extracted
answers using the solution.

\begin{table}[h]
  \caption{\footnotesize{Our solution with an example}}
  \label{tab:solution-example}
  \begin{tabular}{p{.4\linewidth} p{0.55\linewidth}}
    \toprule
    Question: & What are the Amazon RDS storage types?    \\
    \midrule
    Retriever set of documents: & CHAP\_Storage.txt, CHAP\_Limits.txt, CHAP\_BestPractices.txt      \\\\
    Extractor Text Answer: & General Purpose, SSD, Provisioned IOPS, Magnetic    \\\\
    Extractor Yes/No Answer: & None \\
    \bottomrule
  \end{tabular}
\end{table}

\subsection{Retrievers}\label{subsec:retrievers}
Given a question with no context, our approach relies on the retriever to find the right documents
that contains the answer.
The need for a retriever stems from the fact that our extractors are fairly large models and it is time and cost
prohibitive for the extractor to go through all available documents.
For example in our AWS Documentation dataset from Section~\ref{subsec:aws-documentation-dataset}, it will take hours for a
single instance to run an extractor through all available documents.
We ran experiments with simple information retrieval systems with a keyword search along with deep
semantic search models to list relevant documents for a question.
We used precision at K ($P@K$) metric to evaluate our retrievers.
Precision at K is the proportion of retrieved items in the top-k set that are relevant:

\[ P@K = \frac{\text {number of retrieved documents that are relevant}}{\text {total number of retrieved documents}}\]

\subsubsection{Whoosh}\label{subsubsec:whoosh}
Whoosh\footnote{https://whoosh.readthedocs.io/}  is a fast, pure Python search engine library.
The primary design impetus of Whoosh is that it is pure Python and can be used anywhere Python is running,
as no compiler or Java is required.

\subsubsection{Amazon Kendra}\label{subsubsec:amazon-kendra}
Amazon Kendra\footnote{https://aws.amazon.com/kendra/} is a semantic search and question answering service
provided by AWS for enterprise customers.
Kendra allows customers to power natural language-based searches on their own AWS data by using a deep learning-based
semantic search model to return a ranked list of relevant documents.
Amazon Kendra’s ability to understand natural language questions enables it to return the most relevant passage and
related documents.

\subsection{Extractors}
\label{subsec:extractors}
\begin{figure}[h]
  \label{fig:extractor-model}
  \centering
  \includegraphics[width=\linewidth]{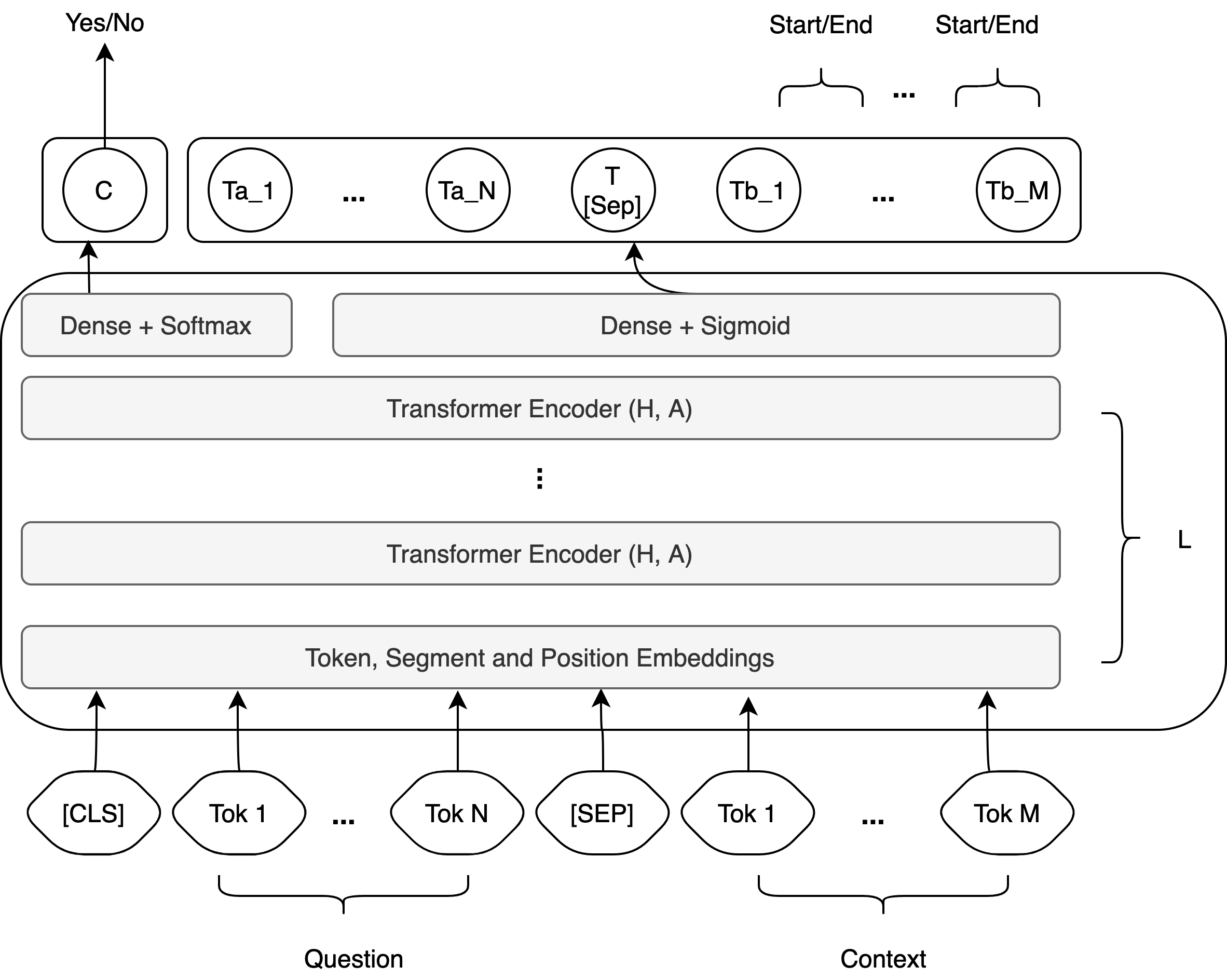}
  \caption{\footnotesize{Extractor model architecture  }}
\end{figure}

Given a question with no context, the retriever finds a set of documents.
Then the output of the retriever will pass on to the extractor to find the right answer for a question.
We created our extractors from a base model which consists of different variations of
BERT~\citep{devlin2018bert} language models and added two sets of layers to extract yes-no-none answers and text answers.
Our approach attempts to find yes-no-none answers and text answers in the same pass.
Our model takes the pooled output from the base BERT model and classifies it in three categories: yes, no, and none.
Furthermore, our model takes the sequence output from the base BERT model and adds two sets of dense layers with
sigmoid as activation.
The first layer tries to find the start of the answer sequences, and the second layer tries to find the end of the
answer sequences.
There can be multiple starts and ends for a single text answer.
Figure~\ref{fig:extractor-model} illustrates the extractor model architecture.

\par
For our base model, we compared BERT (tiny, base, large)~\citep{devlin2018bert} along with RoBERTa~\citep{liu2019roberta},
AlBERT~\citep{lan2019albert}, and distillBERT~\citep{sanh2019distilbert}.
We implemented the same strategy as the original papers to fine-tune these models.
We also used the same hyperparameters as the original papers: L is the number of transformer blocks (layers),
H is the hidden size, and A is the number of self-attention heads.

\subsubsection{Extractor Model}
\label{subsubsec:extractor-model}
We define a training set datapoint as a four-tupele $(d,s,e,yn)$ , where $d$ is a document containing the answers,
 $s, e$ are indices to the start and end of the text answer, and $yn$ defines the yes-no-none answer.
The loss of our model is:
\[ L = \log p (s, e, yn | d) \]
\[ L = \frac{1}{3} (\log ps (s | d) + \log pe (e | d) + \log pyn (yn | d) ) \]

where each probability $p$ is defined as:
\[ ps (s | d) =  \frac {1} {1 + \exp(-f_{start}  (s,d; \theta))} \]
\[ pe (e | d) =  \frac {1} {1 + \exp(-f_{end}  (e,d; \theta))} \]
\[ pyn (yn | d) =  \frac {\exp (-f_{yn}  (yn,d; \theta))} {\sum{yn'} \exp(-f_{yn} (yn',d;\theta))' } \]

where $\theta$ is the base model parameters and $f_{start}$, $f_{end}$, $f_{yn}$ represent three outputs from the
last layer of the model.
\par
At inference, we pass through all text from each document and return all start and end indices with scores higher
than a threshold. We used F1 and Exact Match (EM) metrics to evaluate our extractor models.

\section{Experiments}\label{sec:experiments}
In our experiments, we used pre-trained or ready-to-use information retrieval systems because these systems are
readily available and building a custom retriever with better performance is not economical.
Our experiments show that Amazon Kendra’s semantic search is far superior to a simple keyword search and that
the bigger the base model (BERT-based), the better the performance.
The retriever results are shown in Table~\ref{tab:retrievers}. \par

\begin{table}[h]
  \caption{\footnotesize{Retrievers performance}}
  \label{tab:retrievers}
  \centering
  \begin{tabular}{p{0.15\linewidth} p{0.05\linewidth} p{0.05\linewidth} p{0.05\linewidth} p{0.05\linewidth} p{0.05\linewidth} p{0.05\linewidth} p{0.05\linewidth} p{0.05\linewidth} p{0.05\linewidth} p{0.05\linewidth}}
    \toprule
    Retriever & $P@1$ & $P@3$ & $P@5$ & $P@7$ & $P@9$ & $P@13$ & $P@22$ & $P@30$ & $P@40$ & $P@60$ \\
    \midrule
    Whoosh & .05 & .06 & .06 & .06 & .06 & .06 & .06 & .06 & .06 & .06 \\
    Kendra & .66 & .79 & .86 & .87 & .9 & .91 & .92 & .93 & .94 & .95 \\
    \bottomrule
  \end{tabular}
\end{table}

Regarding our extractors, we initialized our base models with popular pretrained BERT-based models as described in
Section~\ref{subsec:extractors} and fine-tuned models on SQuAD1.1 and SQuAD2.0~\citep{rajpurkar2016squad} along with
natural questions datasets~\citep{kwiatkowski2019natural}.
We trained the models by minimizing loss L from Section~\ref{subsubsec:extractor-model} with the AdamW
optimizer~\citep{devlin2018bert} with a batch size of 8.
Then, we tested our models against the AWS documentation dataset (Section~\ref{subsec:aws-documentation-dataset})
while using Amazon Kendra as the retriever. Our final results are shown in Table~\ref{tab:extractors}.

\begin{table}[h]
  \caption{\footnotesize{Extractor performance}}
  \label{tab:extractors}
  \centering
  \begin{tabular}{p{.3\linewidth} p{0.4\linewidth} p{.1\linewidth} p{.1\linewidth} }
    \toprule
    Extractor Base Model & Extractor Hyperparameters & F1 & EM \\
    \midrule
    BERT Tiny &	L=2, H=128 & 0.128 & 0.09 \\
    RoBERTa Base & L=12, H=768 & 0.154 & 0.09 \\
    DistilBERT & L=12, H=768 & 0.158 & 0.08 \\
    AlBERT Base & L=12, H=768 & 0.199 & 0.11 \\
    BERT Base & L=12, H=768 & 0.245 & 0.16 \\
    BERT Large & L=24, H=1024 & 0.247 & 0.16 \\
    AlBERT XXL & L=12, H=4096 & 0.422 & 0.39 \\
    \bottomrule
  \end{tabular}
\end{table}

\section{Limitations and Future Work}
\label{sec:limitations-and-future-work}
Our solution has a number of limitations.
Below we describe some of these and suggest directions for future work.
We were able to achieve 49\% F1 and 39\% EM for our test dataset due to the challenging nature of
zero-shot open-book problems.
The performance of the solution proposed in this article is fair if tested against technical software documentation.
However, it needs to be improved before finding use in real-world software products.
Additionally, more testing is needed if we want to further expand the applicability of this solution in other domains
(e.g., medical corpus, laws and regulations).
Furthermore, the solution performs better if the answer can be extracted from a continuous
block of text from the document.
The performance drops if the answer is extracted from several different locations in a document.
Moreover, all questions had a clear answer in the AWS documentation dataset,
which is not always the case in the real-world.
As our proposed solution always returns an answer to any question, '
it fails to recognize if a question cannot be answered.\par

For future work, we plan to experiment with generative models such as GPT-2~\citep{radford2019language} and
GPT-3~\citep{brown2020language} with a wider variety of text in pre-training to improve the F1 and EM score
presented in this article.

\section{Conclusion}\label{sec:conclusion}
In this paper, we presented a new solution for zero-shot open-book QA with a two-step architecture to
answer natural language questions from an available set of documents.
With this novel solution, we were able to achieve 49\% F1 and 39\% EM with no domain-specific labeled data.
We hope this new dataset and solution helps researchers create better solutions for zero-shot open-book use cases
in similar real-world environments.

\section*{Declaration of competing interest}\label{sec:Declaration of competing interest}
The authors declare that they have no known competing financial interests or personal relationships
that could have appeared to influence the work reported in this paper.

\section*{CRediT authorship contribution statement}\label{sec:CRediT authorship contribution statement}
\textbf{Sia Gholami}: Conceptualization, Methodology, Software, Investigation, Writing - original draft.
\textbf{Mehdi Noori}: Writing - review \& editing.

\section*{Acknowledgements}\label{sec:Acknowledgements}
The contributions of Sia Gholami and Mehdi Noori were funded by Amazon Web Services.

\vskip 0.2in
\bibliography{zero-shot-open-book-qa}

\end{document}